\documentclass{bmvc2k}

\title{LSTA-Net: Long short-term Spatio-Temporal Aggregation Network for Skeleton-based Action Recognition}

\addauthor{Tailin Chen$^{\ast}$}{t.chen14@newcastle.ac.uk}{1}
\addauthor{Shidong Wang$^{\ast}$}{shidong.wang@newcastle.ac.uk}{2}
\addauthor{Desen Zhou}{zhoudesen@baidu.com}{3}

\addauthor{Yu Guan}{yu.guan@newcastle.ac.uk}{4}

\addinstitution{Open Lab\\
Newcastle University, UK
}

\addinstitution{Open Lab\\
Newcastle University, UK
}

\addinstitution{VIS\\
Baidu, Inc., China
}

\addinstitution{Open Lab \\ Newcastle University, UK
}

\runninghead{T. Chen et al.}{LSTA-Net for skeleton-based action recognition}

\usepackage{multirow} 
\usepackage{bbm}
\usepackage{color}
\usepackage{amssymb}

\begin{document}

\maketitle

\begin{abstract}

Modelling various spatio-temporal dependencies is the key to recognising human actions in skeleton sequences.
Most existing methods excessively relied on the design of traversal rules or graph topologies 
to draw the dependencies of the dynamic joints, which is inadequate to reflect the relationships of the distant yet important joints. 
Furthermore, due to the locally adopted operations, the important long-range temporal information is therefore not well explored in existing works.
To address this issue, in this work we propose LSTA-Net: a novel Long short-term
Spatio-Temporal Aggregation Network, which can 
effectively capture the long/short-range dependencies in a spatio-temporal manner. 
We devise our model into a pure factorised architecture which can alternately perform spatial feature aggregation and temporal feature aggregation. To improve the feature aggregation effect, a channel-wise attention mechanism is also designed and employed.
Extensive experiments were conducted on three public benchmark datasets, and the results suggest that our approach can capture both long-and-short range dependencies in the space and time domain, yielding higher results than other state-of-the-art methods.\footnote[1]{  Code available at \textcolor{magenta}{\url{https://github.com/tailin1009/LSTA-Net}}. $\ast$ Equal Contribution}


\end{abstract}

\section{Introduction}
\label{sec:intro}

In recent years, skeleton-based action recognition 
became of popular research topic in the computer vision
community\cite{aggarwal2014human,st_gcn} due to the advent of cost-effective depth cameras \cite{zhang2012microsoft} and reliable human pose estimation methods\cite{cao2018openpose}. 
Compared with conventional RGB video based action recognition, the data structure of skeleton representation is in low-dimension and hence it can be easily stored in devices and transferred. 

Skeleton-based action recognition is a challenging task due to the lack of context information compared with RGB video based action recognition. 
In particular, modeling the long range dependencies in both spatial and temporal dimensions is difficult due to the flexible configuration between different semantic parts, as well as complex movement patterns in time domain.
Early works used sequential methods to model joint relations, which first extracted features of frames by treating joints as pixels and then utilised RNN-based models to model temporal relationships\cite{du2015hierarchical,liu2016spatio}. some other works concatenated skeletons of different time steps to generate a pseudo image, and utilised CNNs to perform image-based classification\cite{ke2017new,kim2017interpretable}. 
However, human skeletons are not in an Euclidean space and hence above methods cannot model joints' dependencies effectively. 
A more natural method is to model the relationship between joints through a graph neural networks (GNNs). 
In view of the vigorous development of convolution operations, graph convolutional networks (GCNs) have been widely used for skeleton-based action recognition due to the powerful ability for modeling non-Euclidean data. 
In \cite{st_gcn} ST-GCN was proposed, which constructed a spatial graph based on the natural connections of human joints and introduced temporal edges between corresponding joints in consecutive frames. Many of the latest works \cite{li2019spatio,li2019actional,shi2019two,si2019attention,wen2019graph,chen2021learning} can be regarded as variants of ST-GCN, where they typically applied different functional modules for better feature representation. 

	\begin{figure}[t]
		\centering
		\includegraphics[width=0.9\textwidth]{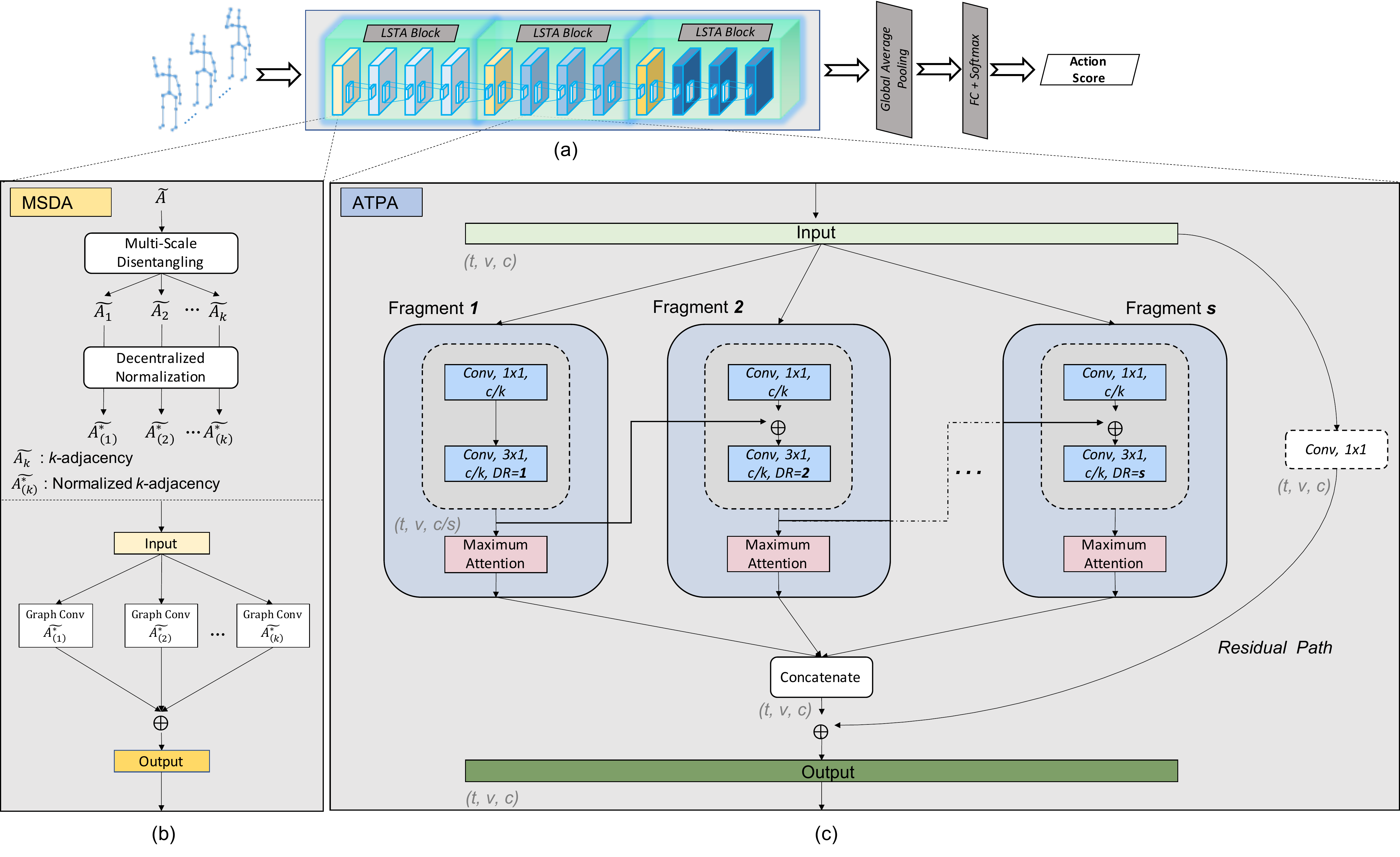}
		\caption{An illustration of the proposed LSTA-Net (a). Each LSTA block includes one MSDA module (b) and three ATPA (c) modules. }
		\label{fig:framework}
		\vspace{-0.2cm}
	    \end{figure}


Despite the significant improvements in performance, there still exists some limitations in the above methods. Specifically, both recurrent and convolutional operations are based on local neighbours in the spatial or temporal domain. 
The capture of long-range dependencies only can be achieved by repeatedly performing these operations and gradually propagating signals through the data and hence it is inefficient. 
Directly modelling the distant joints relations and long-range temporal information is essential for distinguishing various actions. 
Recent work MS-G3D \cite{msg3d} proposed a disentangled and unified spatial-temporal graph convolution strategy to model the long range dependencies in a multi-scale manner. However, the proposed G3D module highly relies on constructing multiple pathways and hence leads to a complex model architecture with high computational costs.

In this paper, to model both long/short-range joints relations in spatial domain, and long/short-term joints dynamics in temporal domain, we propose a novel long short-term spatio-temporal aggregation network (LSTA-Net). 
Specifically, each LSTA block includes a multi-scale decentralised aggregation (MSDA) module and three attention-enhanced temporal pyramid aggregation (ATPA) modules. 
MSDA is proposed to model the semantically related intrinsic connectivity of the disentangled/distant joints in the spatial domain, while ATPA is proposed to model long-range temporal dynamics by 
employing
a set of sub-convolutions and formulating them with a pyramid-like/hierarchical structure.
We further utilise maximum-response attention module (MAM) for further performance improvement. 
The main contributions of our work lie in three folds:
\begin{itemize}
    \item We propose multi-scale spatial decentralised aggregation (MSDA) for distant/long-range dependency modelling by introducing a simple normalisation strategy.
    \item We propose a novel attention-enhanced long short-range temporal modelling architecture, where the temporal receptive field can be efficiently enlarged by the pyramid aggregation scheme. An attention mechanism is also devised for feature enhancement.
    \item Our LSTA-Net, despite smaller model size, achieves higher or comparable results when compared with other state-of-the-arts on three public action recognition datasets, suggesting its effectiveness.
\end{itemize}

\section{Methodology}
In Fig.\ref{fig:framework}, we demonstrate the overall architecture of the proposed LSTA-Net, and in this section we introduce the basic unit LSTA block.  
Each LSTA block contains a multi-scale spatial decentralised aggregation module (MSDA , i.e., Fig.\ref{fig:framework}(b)), and three attention-enhanced temporal pyramid aggregation (ATPA, i.e.,Fig.\ref{fig:framework}(c)) modules to extract the spatial and temporal features for skeleton-based action recognition. 
    

	\subsection{Multi-scale Spatial Decentralised Aggregation(MSDA)}
	
	\subsubsection{Preliminaries}
	\noindent\textbf{Notations} A graph is defined as $\mathcal{G}=(\mathcal{V}, \mathcal{E})$, where $\mathcal{V}$ is the set of vertices and $\mathcal{E} \subseteq (\mathcal{V} \times \mathcal{V})$ is the set of edges. 
	The relationship of the graph is represented by the graph adjacency matrix $\mathbf{A} \in \mathbb{R}^{V \times V}$ with entry $\mathbf{A}_{ij}=1$, when nodes $i$, $j$ are connected, and 0 otherwise; \textit{V} denotes the number of vertexes.
	$\mathbf{A}$ is a symmetric matrix while $\mathcal{G}$ is an undirected graph. Human graph sequences contain a set of node features $\mathcal{X}=\{{x}_{t}^{v}|1\leq v \leq V, 1\leq t \leq T; v, t \in \mathbb{Z}\}$
	that can also be represented as
	$\textbf{X} \in \mathbb{R}^{T \times V \times C}$, where $C$ is the feature dimension.
	
	
	

	\noindent\textbf{Multi-Scale Aggregation}
	The multi-scale spatial aggregation \cite{li2019actional} on a given graph can be implemented similar to the convolution on a regular grid graph, such as the RGB image. 
	At timestamp $t$, given input graph feature $\textbf{X}_{t}\in \mathbb{R}^{V \times C}$, via a multi-scale GCN operation we can get the output skeleton feature 
	\begin{equation}\label{eq1}
	\textbf{X}_{t}^{\rm{spat}} = \sigma \left( \sum_{k=0}^{K} \widetilde{\textbf{A}_{k}} \textbf{X}_{t} \textbf{W}_{k} \right),
	\end{equation}
	where \textit{K} is the number of scales of the graph to be aggregated; 
	$\sigma(\cdot)$ is the activation function; $\tilde{\textbf{A}}$ is the normalised adjacency matrix \cite{kipf2016semi} that can be obtained by: $\tilde{\textbf{A}}=\hat{\textbf{D}}^{-\frac{1}{2}}\hat{\textbf{A}}\hat{\textbf{D}}^{-\frac{1}{2}}$, where $\hat{\textbf{A}}=\textbf{A}+\textbf{I}$,
	$\textbf{I}$ and $\hat{\textbf{D}}$ are the identity matrix and degree matrix of \textbf{A} respectively.   
	The term $ \widetilde{\textbf{A}_{k}} \textbf{X}_{t} $ in Eq.(\ref{eq1}) describes a weighted sum feature that is based on the \textit{k}-order neighbourhood of the selected graph nodes.

	\subsubsection{Decentralised Normalisation Strategy for MSDA}
	One drawback of the aforementioned multi-scale aggregation approaches is that it may overweight the low-order neighbours \cite{li2019actional,ms-aagcn}.
	To solve such biased weighting problem, in \cite{msg3d} Liu et al. proposed a disentangled aggregation scheme. However, such disentangled representation neglected the correlations between different scales and hence suffered from limited spatial representation capacity.
   To address this issue, here we employ a simple decentralised normalisation strategy based on the disentangled representation which aims to expand such representation to multiple scales. 
	
	In the decentralised normalisation strategy, the elements of \textit{k}-adjacency matrix $ \widetilde{\textbf{A}_{(k)}} $ are first assigned the value 1 if node $ i = j $ or with their shortest distance $ d(i, j) =k$. For those elements whose shortest distance $ d(i, j) < k$, a scale-adaptive value $  \frac{d(i, j)}{k}$ is assigned. The rest of elements are set to 0. 
	

	
	It is clear when $k=0$ or $k=1$, we have $\widetilde{\textbf{A}_{(0)}}=\textbf{I}$, or $\widetilde{\textbf{A}_{(1)}}=\widetilde{\textbf{A}}$. 
	When $k\ge2$,  similar to \cite{msg3d}, the normalised \textit{k}-adjacency matrix $\widetilde{\textbf{A}^*_{(k)}}$  can be obtained by calculating the residuals of the matrix powers of current graph scale and the mean of previous graph scales, i.e.,  
	\begin{equation}\label{decentralized}
	\widetilde{\textbf{A}^*_{(k)}} = \textbf{I} + \mathbbm{1} \left(\widetilde{\textbf{A}_{k}}\geq 1\right) -\mathbbm{1} \left( ({ \frac{1}{k}} {\sum ^{k-1}_{n=0}}{\widetilde{\textbf{A}_{n}}})\geq 1 \right).
	\end{equation}	  
	By substituting $\widetilde{\textbf{A}_{k}}$ with $\widetilde{\textbf{A}^{*}_{(k)}}$ in Eq.(\ref{eq1}), spatial feature can be extracted via: 
	\begin{equation}\label{eq3}
		\textbf{X}_{t}^{\rm{spat}} = \sigma \left( \sum_{k=0}^{K} \widetilde{\textbf{A}^{*}_{(k)}} \textbf{X}_{t} \textbf{W}_{(k)} \right),
	\end{equation}  
	which is our MSDA module. 
	Compared with \cite{msg3d}, the application of decentralised normalisation strategy (i.e., with the scale-adaptive value assigning scheme) in MSDA can well summarise multiple scales, capturing the dependencies between short and long-range joints in the spatial domain.
	

	\subsection{Attention-enhanced Temporal Pyramid Aggregation(ATPA)}
	
MSDA is able to capture the dependencies between both short and long-range joints in space domain, and it is desirable to investing the long short-range temporal modeling.
Here the proposed Temporal Pyramid Aggregation (TPA) module divides the convolution operation of the input features into a group of subsets, 
which can effectively expand the equivalent receptive field of the temporal dimension without introducing additional parameters or time-consuming operations. 
	 
	 

	 
	 \subsubsection{Temporal Pyramid Aggregation  (TPA)}
	After spatial feature extraction (using MSDA, i.e., Eq.(\ref{eq3})), the original skeleton graph sequence can be represented as $ \textbf{X}^{\rm{spat}} \in \mathbb{R}^{T \times V \times C'}$, where $C'$ is the feature dimension. 
	In this subsection, we introduce temporal pyramid aggregation (TPA) for effective temporal information encoding. 
	
	To exploit the temporal information in skeleton-based action recognition, previous works \cite{st_gcn,shi2019two,li2019spatio,li2019actional} used temporal convolution on the neighbouring timestamps/frames, and performed repeated stacking for long-range temporal dependency modelling. 
	However, useful features from distant frames may have been weakened after a large number of local convolution operations. 
	In \cite{msg3d}, Liu et al. expanded the temporal receptive field by composing a large number of local operations, which increase the model size substantially, with extremely high computational costs. 
	In \cite{res2net,zhang2020resnest}, Res2Net-like architectures were introduced, 
	which deformed the ordinary convolution layer into a set of sub-convolutions and constructed hierarchical residual-like connections to capture multi-scale feature representations. 
	Motivated by \cite{res2net,zhang2020resnest}, here we employ this concept to skeleton-based action recognition for fast and efficient long-range temporal dependencies modelling. 
	
	Given $ \textbf{X}^{\rm{spat}} \in \mathbb{R}^{T \times V \times C'}$, along the feature dimension, we can obtain $S$ embedded fragments $\{\textbf{X}_{s}|s = 1,2,...,S; \textbf{X}_{s} \in \mathbb{R}^{T \times V \times \alpha}\}$ through multiple learnable transformations, where $\alpha$ is the feature dimension in the embedding space, and we set $\alpha = \lfloor C'/S \rfloor$. 
    Then, these fragments can be formulated as a hierarchical residual architecture and thus can be hierarchically processed by temporal convolutions with gradually increasing dilation rates. 
	Specifically, assume $S = 6$, this process can be written as: 
	\begin{equation}\label{eq4}
	\begin{aligned}
	&\textbf{X}_{s}^{\rm{temp}}=conv_{\rm{temp}}*\textbf{X}_{s}, \qquad \qquad \qquad \qquad \quad \ s=1,\\
	&\textbf{X}_{s}^{\rm{temp}}=conv_{\rm{temp}}*\left(\textbf{X}_{s}+\textbf{X}_{s-1}^{temp}\right), \qquad s=2,3,4,5,6,
	\end{aligned}
	\end{equation}     
	where $ \textbf{X}_{s}^{\rm{temp}} $ is the output of temporal convolution in \textit{s}-th fragment; $ conv_{\rm{temp}} $ denotes the $3 \times 1$ temporal sub-convolution. 
	
	The above operations endow the different fragments with different receptive fields in temporal dimension, which can model the long-range temporal dependencies effectively. 
	The final output can be easily obtained by concatenating outputs of multiple temporal convolutions as follows:
	\begin{equation}\label{eq5}
	\textbf{X}^{\rm{temp}}=[\textbf{X}_{1}^{\rm{temp}};\textbf{X}_{2}^{\rm{temp}};\textbf{X}_{3}^{\rm{temp}};\textbf{X}_{4}^{\rm{temp}};\textbf{X}_{5}^{\rm{temp}};\textbf{X}_{6}^{\rm{temp}}].
	\end{equation}
	Through TPA, we can obtain representation $ \textbf{X}^{\rm{temp}}$, which has encoded various range of temporal information. 
	
	\subsubsection{Maximum-response Attention Module (MAM)} 
\begin{figure}[!tb]
		\centering
		\includegraphics[width=0.4\textwidth]{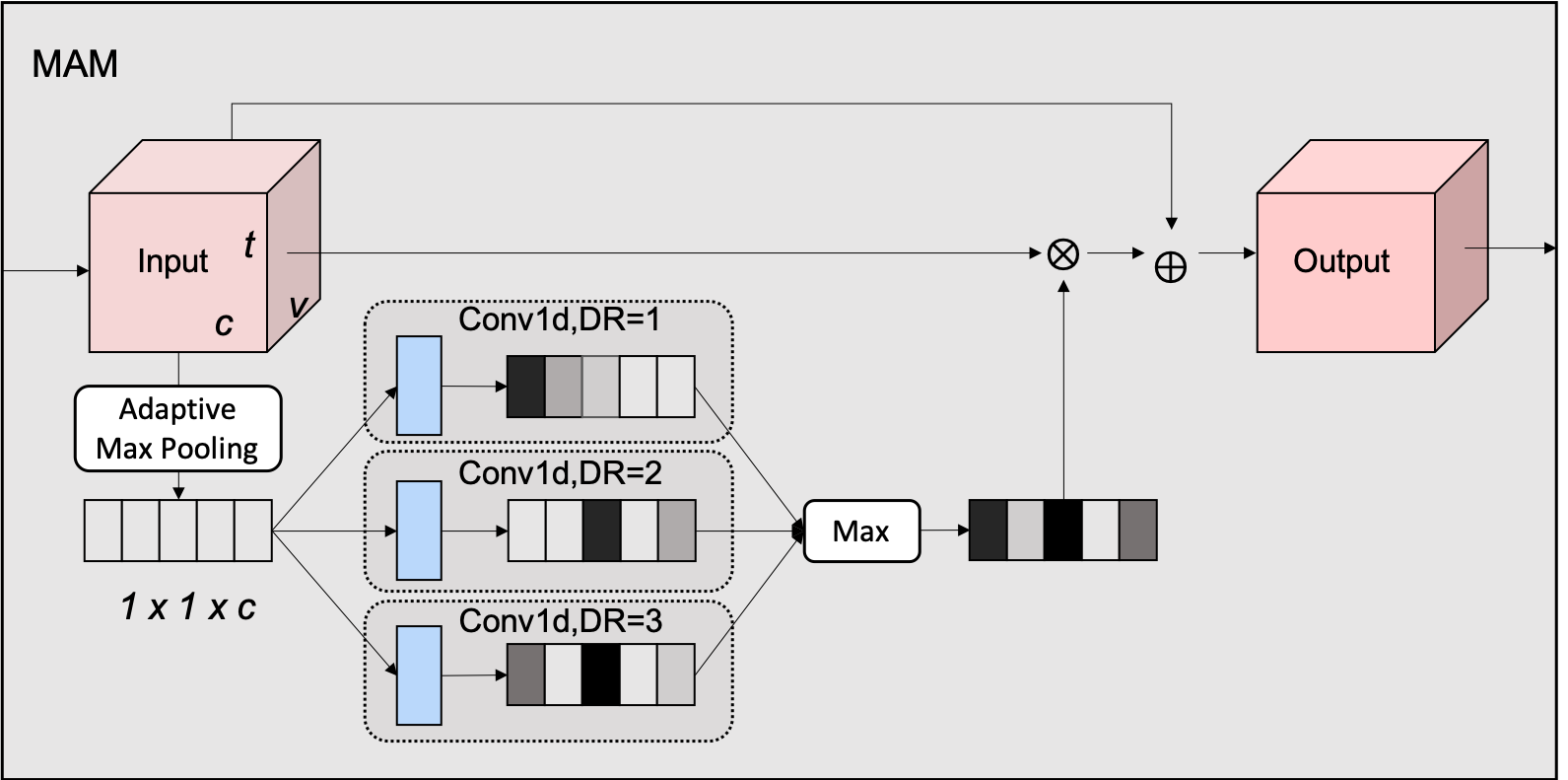}
		\caption{An illustration of the proposed MAM Module. }
		\label{fig:MAM}
		\vspace{-0.5cm}
    \end{figure}

	Although the combination of MSDA and TPA modules can model such long-range and short-range spatial temporal dependencies, human skeleton sequences comprise a limited number of dynamic key joints, which supplies a favourable breeding ground for the development of attention mechanisms. Employing attention mechanism can spontaneously capture useful intrinsic correlations without knowing the content of the input sequence. These motivate us to explore an effective and efficient attention mechanism to extract the semantic dependence of skeletal data. Very few works explored the attention mechanism in skeleton-based action recognition. In \cite{ms-aagcn}, Shi et al. applied SE-like\cite{hu2018squeeze} attention modules to re-weight the feature maps in spatial, temporal and channel dimension sequentially, yet it failed in capturing the joint attention. In \cite{eca}, an Efficient Channel Attention (ECA) module was used to capture channel attention, yet it only included a fixed kernel and may be inadequate when facing complex data. 
    Motivated by  \cite{eca} in our multi-scale aggregation schemes, we expand the original ECA from 
single path to multiple paths and then use an adaptive maximum pooling operation for the most important features. 
	In Fig.\ref{fig:MAM}, we can see the structure of this Maximum-response Attention Module(MAM). 
	Given an input  
	$ \textbf{X} \in \mathbb{R}^{ T \times V \times C} $, 
the channel attention can be computed by a standard 1\textit{D} convolution with kernel size of $ \eta $ :

	
	\begin{equation} \label{eq9}
		\omega = \tau ( conv1D_{ \eta }(g( \textbf{X} ))).
	\end{equation}

where $\tau(\cdot)$ is the sigmoid function; $g(\textbf{X})$ denotes the function of 2D Adaptive Max Pooling operated on spatial and temporal dimensions. 
	Although the 1\textit{D} convolution only has $ \eta $ parameters, it
	provides a very limited receptive field, which leads to insufficient local information interaction. 
	To tackle this problem, we extend the 1\textit{D} convolution to a parallel fashion with different dilation rates. Subsequently, we apply the element-wise maximum operator to obtain the highest response of the input feature to the classifier. We propose to formulate this manipulation in the following manner:
	\begin{equation}\label{eq10}
	\omega_{max}=\max_{\phi \in \Phi}\left(\phi(\omega) \right).
	\end{equation}
	where $\phi(\omega)$ represents a set of 1\textit{D} convolutions stacking with different dilation rates, and $\Phi$ is the total length of attending convolutions. The new features $ \omega_{max} $ is constructed in such a way not only enhances the interactivity of information, but also makes their output independent of the receptive fields known in advance by the various kernels.  
    
\subsection{LSTA Block} \label{model_architecture}
For skeleton-based action recognition, the joint number tends to be small (e.g., 25) when compared with the frame number (e.g., 300). 
Therefore, we use more ATPA modules (for temporal modelling) than MSDA modules (for spatial modelling) for each LSTA block. 
Specifically in our LSTA-Net, each LSTA block is composed of  
one MSDA module followed by three ATPA modules to perform spatial temporal aggregation:
    	\begin{equation}\label{eq11}
            M_{\rm{ATPA}}(M_{\rm{ATPA}}(M_{\rm{ATPA}}(M_{\rm{MSDA}}(\textbf{X})))),
    	\end{equation}
where $\textbf{X}$ denotes the input.   
In our LSTA-Net, we set multiple LSTA blocks as shown in Fig. \ref{fig:framework}, and more details can be found in Experiments Section.    	

	\begin{table*}[ht!]
		\centering
		\scalebox{0.86}{
			\begin{tabular}{l|c|cc|cc}
				\hline
				\multirow{2}{*}{Methods} & 
				 \multirow{2}{*}{Params} & \multicolumn{2}{c}{NTU RGB+D 60} & \multicolumn{2}{c}{NTU RGB+D 120} 
				\\ \cline{3-4} \cline{5-6}  & & X-Sub & X-View & X-Sub & X-Set    
				\\  \hline \hline

				PA-LSTM \cite{ntu_60}                       & -- &  60.7 & 67.3 & 25.5  & 26.3    \\
			    ST-LSTM \cite{liu2016spatio}               & -- &  69.2 & 77.7 & 55.0  & 57.9    \\
				 VA-LSTM \cite{zhang2017view}               & -- &  79.4 & 87.6 & --    & --      \\
				TCN \cite{kim2017interpretable}            & -- & 74.3 & 83.1 & --    & --      \\
				
				AGC-LSTM \cite{si2019attention}            & -- &  89.2 & 95.0 & --    & --      \\ 
				\hline
				ST-GCN \cite{st_gcn}                    & 3.1M &  81.5 & 88.3 & 70.7  & 73.2    \\
				AS-GCN \cite{li2019actional}               & - &  86.8 & 94.2 & 77.9  & 78.5    \\
				2s-AGCN \cite{shi2019two}                  & 6.9M &  88.5 & 95.1 & 82.9  & 84.9    \\
				DGNN \cite{shi2019skeleton}                 & 26.2M &  89.9 & 96.1 & --    & --      \\
				NAS-GCN \cite{peng2020learning}          & 6.6M &  89.4 & 95.7 & --    & --      \\
				4s-Shift-GCN \cite{cheng2020skeleton}      & 2.8M &  90.7 & 96.5 & 85.9  & 87.6    \\
				4s-DC-GCN+ADG \cite{decouple_gcn_cheng2020eccv}     & - &  90.8 & \textbf{96.6} & 86.5  & 88.1    \\

                PA-ResGCN \cite{ResGCN}                     &  3.6M & 90.9 & 96.0 & 87.3  & 88.3    \\
				MS-G3D \cite{msg3d}                         & 6.4M &  \textbf{91.5} & 96.2 & 86.9 & 88.2 \\
    			\hline

				\textbf{LSTA-Net(Ours)}   & 3.1M & \textbf{91.5} & \textbf{96.6} & \textbf{87.5} & \textbf{89.0}  \\
                \hline
		\end{tabular}}
		\caption{Model Comparison (in Top-1 accuracy (\%)) on the NTU RGB+D 60 \& 120 datasets.}
		\label{ntu60,120}
	\end{table*}

\section{Experiments}
	
	\subsection{Datasets}
	
    \noindent\textbf{NTU RGB+D 60} \textbf{and} \textbf{NTU RGB+D 120} NTU RGB+D 60\cite{ntu_60} and 120\cite{ntu_120} contain 56,580 and 114,480 skeleton sequences corresponding to 60 and 120 action categories respectively. 
    Each skeleton sequence contains the 3D spatial coordinate information of 25 joints captured by Microsoft Kinect v2 cameras. 
    We follow the same protocols in prior works\cite{shi2019two,msg3d},i.e., 
    Cross-Subject(X-Sub) and  Cross-View (X-View) experiments for NTU RGB+D 60; while Cross-Subject(X-sub) and  Cross-Subject(X-sub) for  NTU RGB+D 120;
    More details of the train/test splits in different settings can be found in \cite{shi2019two,msg3d}.
    
    
    \noindent\textbf{Kinetics-Skeleton} The Kinetics-Skeleton is a large-scale dataset sourced from the Kinetics 400 video dataset \cite{kinetics} by using the OpenPose\cite{cao2018openpose} pose estimation toolbox. Effective skeleton sequences are divided into two groups, 240,436 for training and 19,796 for testing. Each skeleton sequence contains the 2D spatial coordinate information of 18 joints and the corresponding confidence scores. Following the previous works\cite{msg3d,shi2019two}, Top-1 and Top-5 accuracies are reported.

	\subsection{Implementation Details}
	We implemented the proposed LSTA-Net using PyTorch toolkit and ran on a server with four Tesla-V100 GPUs.
	The batch size was set to 64 (16 per worker). The model was trained for 100 epochs with Nesterov momentum (0.9) SGD and the cross-entropy loss. The initial learning rate was set to 0.05 and decayed with a factor of 0.1 at epoch \{40,60,80,100\}. The weight decay was set to 0.0005 for all experiments. The architecture is the same as the factorised path in \cite{msg3d} but with the different output channels (i.e., 72, 144 and 288 for each LSTA block in sequential). 
	The input skeletal data is padded to \textit{T}=300 frames by replaying the actions. All sequences were pre-processed with normalisation and translation as employed in \cite{shi2019skeleton,shi2019two,msg3d}.
	
		\subsection{Comparison with State-of-the-arts}

	\begin{table}[t!]
		\centering
		\scalebox{0.86}{
			\begin{tabular}{lccc}
				\hline

				\multirow{2}{*}{Methods}  &\multicolumn{2}{c}{Kinetics Skeleton} \\ \cline{2-3} & Top-1     & Top-5     \\ 
				\hline \hline
				
				PA-LSTM \cite{ntu_60}            & 16.4           & 35.3            \\ 
				TCN \cite{kim2017interpretable}	 & 20.3	&40.0\\
				ST-GCN \cite{st_gcn}             & 30.7           & 52.8            \\
				AS-GCN \cite{li2019actional}      & 34.8           & 56.5            \\
				2s-AGCN \cite{shi2019two}         & 36.1           & 58.7            \\
				DGNN \cite{shi2019skeleton}       & 36.9           & 59.6            \\
				NAS-GCN \cite{peng2020learning}   & 37.1           & 60.1            \\
				MS-G3D \cite{msg3d}               &   38.0 & \textbf{60.9} \\     
				
				\hline

				\textbf{LSTA-Net (Ours)}            &  \textbf{38.1} & 60.7   		\\
				\hline

		\end{tabular}}
		\caption{Model Comparison (in Top-1/5 accuracy (\%)) on the Kinetics Skeleton dataset.}
		\label{kinetics}
	\end{table}

{\color{blue}
	\begin{table}[t!]
	    \centering
	    \scalebox{0.86}{
		\begin{tabular}{lccc}
			\hline
			Method & Params & X-Sub & X-View  \\
			\hline \hline
			Shift-GCN\cite{cheng2020skeleton} & 0.7M & 87.8 & 95.1 \\
			MS-G3D\cite{msg3d}  & 3.2M  &  \textbf{89.4} &95.0    \\
			\hline
			\textbf{LSTA-Net(Ours)}  & 1.0M  & 89.1  & \textbf{95.3}    \\ 
			\hline
		\end{tabular}}
		\caption{Model comparison (in parameter number and Top-1 accuracy(\%)) on the NTU RGB+D 60 dataset (joint data only).}
		\label{single modality comparison}
	\end{table}

	}

    
    Many state-of-the-art(SOTA) methods utilised multi-stream fusion strategies to fuse different modalities data for higher results. 
    For fair comparison, we employed the similar multi-stream fusion strategy as \cite{cheng2020skeleton,shi2019skeleton} , and devised our framework in the three-stream fashion where joint, bone, motion streams were sharing one identical architecture. 
    The initialisation of the "bone" stream was set to the vector difference of adjacent joints directed away from the center of the human body. Then the "motion" stream used the temporal difference between adjacent frames of "joint" or "bone" as input. Finally, a score-level fusion strategy was applied to obtain the final prediction score.

	We compare our full model with other state-of-the-art methods on NTU-RGB+D 60,  NTU-RGB+D 120 and Kinetics-Skeleton datasets and the results are shown in Table \ref{ntu60,120} and Table \ref{kinetics}. On NTU RGB+D 60 dataset, we achieve competitive performance on cross-view and cross-subject benchmarks. For NTU RGB+D 120, our method outperforms other methods on both cross-subject and cross-setup benchmarks. 
	We additionally compare model parameter number with several SOTA GCN models in Table \ref{ntu60,120}, suggesting our proposed LSTA-Net is a light-weight scheme yet with the best performance on both datasets.
	For Kinetics-skeleton dataset, as can be seen from Table \ref{kinetics}, our proposed LSTA-Net achieves comparable performance with MS-G3D \cite{msg3d} and outperforms others. 
	However, our model only has $50\%$ of the parameters as in MS-G3D\cite{msg3d}, suggesting the effectiveness of our aggregation scheme. 

In Table \ref{single modality comparison}, we report the  experimental results using only the original skeleton data on NTU RGB+D 60 dataset, and compare with other approaches (i.e., Shift-GCN\cite{cheng2020skeleton} and MS-G3D\cite{msg3d}).
Although shift-GCN\cite{cheng2020skeleton} is a light-weight model with only 0.7M parameters, 
our model outperforms it by a large margin with only 0.3M additional parameters, suggesting it is an alternative solution at balancing the trade-off between efficiency and effectiveness. 
On the other hand, when compared with MS-G3D\cite{msg3d}, our method achieves comparable results with only 1/3 parameters. 
These observations suggest our approach is an effective light-weight solution. 

\subsection{Ablation Study}
In this section, we report the results of our ablation study to validate the effectiveness of our proposed model components or strategies. 
Unless stated, performance is reported as classification accuracy on the Cross-Subject benchmark of NTU RGB+D 60 dataset using the joint data only.
    
	\begin{table}[!t]
	    \centering
	    \scalebox{0.86}{
		\begin{tabular}{lcccc}
			\hline
			\multirow{2}{*}{Spatial Aggregation} & \multicolumn{4}{c}{Number of Scales} \\ \cline{2-5} & k=1     & k=4     & k=8     & k=12                                                                                \\ \hline\hline
		    GCN       & 87.1  & 88.2    & 88.6    & 88.1  \\ 
		    MS-GCN\cite{msg3d}   & 87.1  & 88.2     & 88.9    & 88.2  \\
		    MSDA (Ours) & 87.1  & 88.3   & \textbf{89.1}    & 88.3  \\
        \hline
		\end{tabular}}
		\caption{Ablation study on spatial aggregation schemes, with top-1 accuracies reported (\%).}
		\label{Ablation_MSDA}
	\end{table}

\noindent\textbf{MSDA module}
In Table \ref{Ablation_MSDA}, we compare the proposed MSDA with the basic adjacency powering method and disentangling \cite{msg3d} method in terms of scale number. 
We replace the spatial aggregation strategy of the LSTA blocks, referred to as "GCN", "MS-GCN"\cite{msg3d} and "MSDA", respectively. We observe that our decentralised aggregation strategy MSDA can outperform basic adjacency powering method on different scales. 

	\begin{table}[t!]
	    \centering
	    \scalebox{0.86}{
		\begin{tabular}{lccc}
			\hline
			Temporal Aggregation & Params & Attention & Acc  \\
			\hline \hline
			MS-TCN\cite{msg3d}  & 1.2M  & -  &88.2    \\ 
			TPA(Ours)           & 1.0M  & -   &88.5      \\
			ATPA(Ours)           & 1.0M  & \checkmark  &\textbf{89.1}    \\
			\hline
		\end{tabular}}
		\caption{Comparisons between regular MS-TCN and our TPA module with or without Temporal Maximum Attention, with Top-1 accuracies reported (\%).}
		\label{compare_with_ms_tcn}
	\end{table}

\noindent\textbf{ATPA module}
To validate the effectiveness of our attention-enhanced temporal aggregation method, we conducted ablation experiments on different temporal aggregation schemes, and the results are shown in Table \ref{compare_with_ms_tcn}. From the table we can see that our proposed TPA (w/o attention) scheme outperforms the direct aggregation  MS-TCN\cite{msg3d}. The final result can be further improved when combining with maximum response attention (ATPA). We also conducted extensive experiments to explore the hyper-parameters $S$ in TPA and $\eta$ in MAM module, and the results are shown in Table \ref{TPA_differnet_S} and Table \ref{mam_ablation_study}, suggesting both hyper-parameters are quite stable. 
Additionally, we compare different pooling functions in our MAM module. As shown in Table \ref{mam_ablation_study}, the max-pooling yields the best accuracy while the average-pooling achieves more stable results, indicating it is less sensitive to kernel size. 
Furthermore, at the end of the entire network, a global average pooling is adopted for final feature generation.

	\begin{table}[t!]
	    \centering
	    \scalebox{0.86}{
		\begin{tabular}{ccc}
		
			\hline
			 Method  & Number of Subsets & Acc \\
			    \hline \hline
             \multirow{3}{*}{ATPA}    &  $S=4$ &  89.0   \\ 
                                     & $S=6$ &  \textbf{89.1}   \\
                                     & $S=8$ &  88.9   \\
			\hline
		\end{tabular}}
		\caption{Comparisons between TPA modules with respect to $S$. $S$ is the number of subsets for sub-convolution operations, with Top-1 accuracies reported (\%).}
		\label{TPA_differnet_S}
		\vspace{-0.4cm}
	\end{table}

	\begin{table}[t!]
	   \centering
		\begin{tabular}{lc|cccc|c|c}
			\hline
			\multirow{3}{*}{MAM} & \multicolumn{7}{c}{Number of Dilation Rates}              \\ \cline{2-8}
			& $\eta=3$      & \multicolumn{4}{c}{$\eta$=5}            & $\eta$=7      & $\eta$=9    \\	\cline{2-8}
			& 1$\sim$3 & 1 & 1$\sim$2 & 1$\sim$3 & 1$\sim$4 & 1$\sim$3 & 1$\sim$3 \\
			\hline \hline 
         w/ Average-Pooling &88.5 &88.3   &88.5 &88.8 &88.5 &88.3 &88.2        \\
		w/ Max-Pooling &88.6 &88.3 &88.6 &\textbf{89.1} &88.4 &88.4 &88.1        \\
			
			\hline 
		\end{tabular}
	\caption{Parameter selection of MAM { with Average-Pooling/Max-Pooling functions}. $\eta$ denotes the kernel size of 1\textit{D} convolution as in Eq.(\ref{eq9}), and 1$\sim$3 indicates the dilation rates of 1, 2 and 3.}
	\label{mam_ablation_study}
	\vspace{-0.1cm}
	\end{table}


	
\begin{figure}[!tb]
		\centering
		\includegraphics[width=1\textwidth]{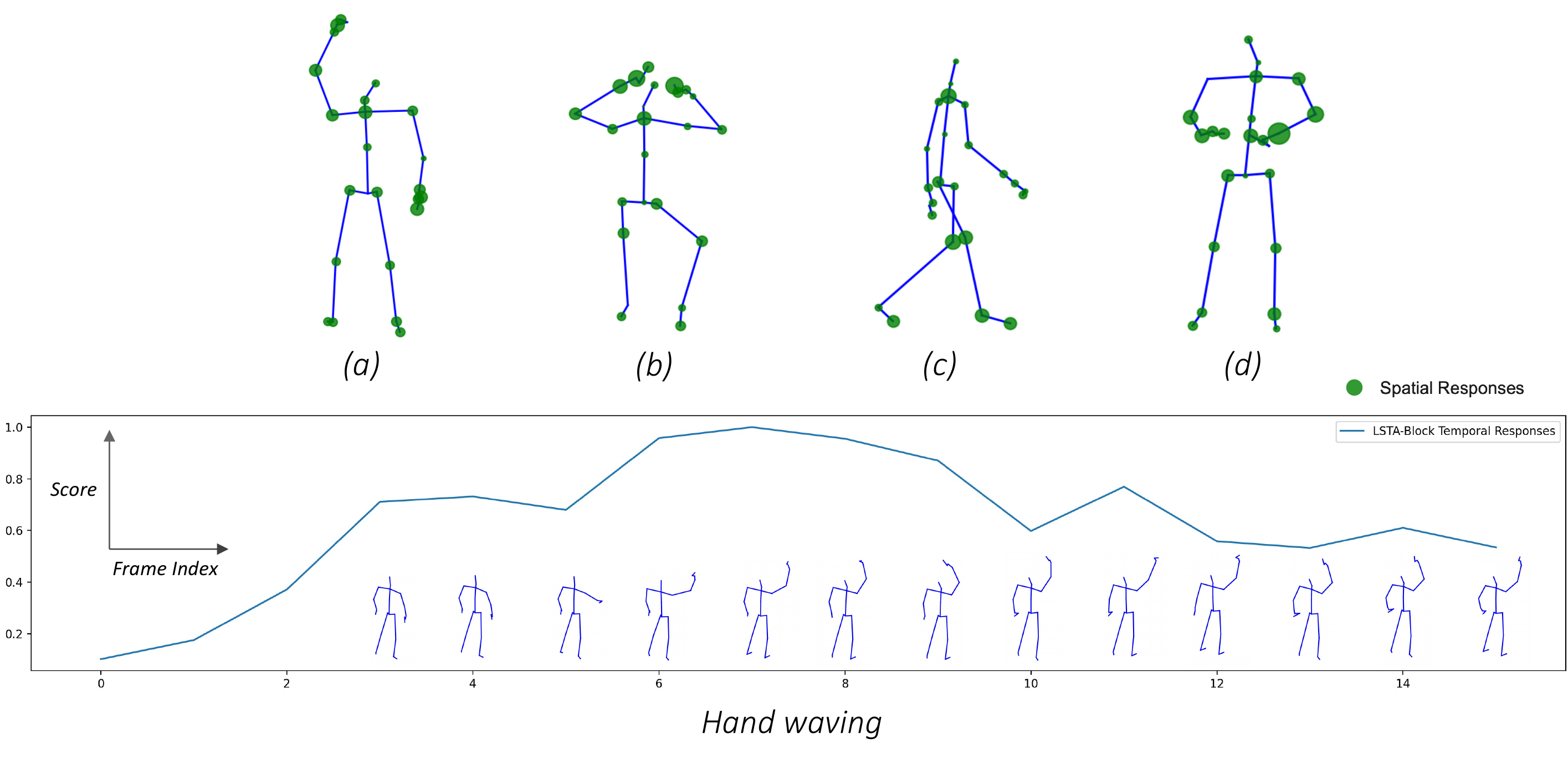}
		\caption{TOP: Examples of the joint feature responses for four actions (a) "walking" (b) "put on a hat" (c)"hand waving" (d) "type on a keyboard". {The size of green circles indicates the importance of the joint}.  
		{BOTTOM:} Visualisation of the temporal feature responses for each of the frame for action "hand waving". {X-axis denotes the input skeleton frame index, and Y-axis indicates the importance of each frame (scaled to range $[0,1]$). 
		}.
		}
		\label{fig:Spatial_Temporal_Feature_Responses}
		\vspace{-0.4cm}
    \end{figure}

\noindent\textbf{Visualisation}
We visualise the output feature maps of the last LSTA-block in Fig.\ref{fig:Spatial_Temporal_Feature_Responses}. 
For the spatial modelling (TOP), the size of green circle around each joint indicates its importance. 
We can see our model can focus on the parts that are most relevant to the action. Specifically, both hands are well focused for actions "type on a keyboard", "put on a hat"; the model focuses more on arm parts for action "hand waving"; for "walking" action, the model focuses on the lower body, especially the feet and knees. 

For the temporal modelling, we show an example of the learned feature responses for several frames and the corresponding skeleton sketches are in Fig.\ref{fig:Spatial_Temporal_Feature_Responses} (BOTTOM). 
For action "hand waving", we can see the model focuses more on the process of raising hand in temporal domain and also pays attention on localised motion patterns in the spatial domain (i.e., "the raised hand"), suggesting the capability of our model in capturing the spatio-temporal dependencies in skeleton-based action recognition. 

    



\section{Conclusion}
In this work, we propose the light-weight LSTA-Net, which can alternately performs long short-term spatio-temporal feature aggregation for improved skeleton-based action recognition. 
Specifically, MSDA for capturing distant spatial information, and ATPA for capturing long-range temporal information are proposed, with MAM employed for further performance gain. 
On three large-scale public datasets, despite smaller model size, our LSTA-Net achieves higher accuracies than most of other state-of-the-arts, suggesting it is a practical solution for skeleton-based action recognition. 

\section{Acknowledgment}
This research is jointly funded by EPSRC Centre for Doctoral Training in Digital Civics (EP/L016176/1) and EPSRC DERC: Digital Economy Research Centre (EP/M023001/1).

\bibliography{egbib}

\begin{thebibliography}{30}
\providecommand{\natexlab}[1]{#1}
\providecommand{\url}[1]{\texttt{#1}}
\expandafter\ifx\csname urlstyle\endcsname\relax
  \providecommand{\doi}[1]{doi: #1}\else
  \providecommand{\doi}{doi: \begingroup \urlstyle{rm}\Url}\fi

\bibitem[Aggarwal and Xia(2014)]{aggarwal2014human}
Jake~K Aggarwal and Lu~Xia.
\newblock Human activity recognition from 3d data: A review.
\newblock \emph{Pattern Recognition Letters}, 2014.

\bibitem[Cao et~al.(2018)Cao, Hidalgo, Simon, Wei, and Sheikh]{cao2018openpose}
Zhe Cao, Gines Hidalgo, Tomas Simon, Shih-En Wei, and Yaser Sheikh.
\newblock Openpose: realtime multi-person 2d pose estimation using part
  affinity fields.
\newblock \emph{arXiv preprint arXiv:1812.08008}, 2018.

\bibitem[Chen et~al.(2021)Chen, Zhou, Wang, Wang, Guan, He, and
  Ding]{chen2021learning}
Tailin Chen, Desen Zhou, Jian Wang, Shidong Wang, Yu~Guan, Xuming He, and Errui
  Ding.
\newblock Learning multi-granular spatio-temporal graph network for
  skeleton-based action recognition.
\newblock In \emph{ACMMM}, 2021.

\bibitem[Cheng et~al.(2020{\natexlab{a}})Cheng, Zhang, Cao, Shi, Cheng, and
  Lu]{decouple_gcn_cheng2020eccv}
Ke~Cheng, Yifan Zhang, Congqi Cao, Lei Shi, Jian Cheng, and Hanqing Lu.
\newblock Decoupling gcn with dropgraph module for skeleton-based action
  recognition.
\newblock In \emph{ECCV}, 2020{\natexlab{a}}.

\bibitem[Cheng et~al.(2020{\natexlab{b}})Cheng, Zhang, He, Chen, Cheng, and
  Lu]{cheng2020skeleton}
Ke~Cheng, Yifan Zhang, Xiangyu He, Weihan Chen, Jian Cheng, and Hanqing Lu.
\newblock Skeleton-based action recognition with shift graph convolutional
  network.
\newblock In \emph{CVPR}, 2020{\natexlab{b}}.

\bibitem[Du et~al.(2015)Du, Wang, and Wang]{du2015hierarchical}
Yong Du, Wei Wang, and Liang Wang.
\newblock Hierarchical recurrent neural network for skeleton based action
  recognition.
\newblock In \emph{CVPR}, 2015.

\bibitem[Gao et~al.(2019)Gao, Cheng, Zhao, Zhang, Yang, and Torr]{res2net}
Shanghua Gao, Ming-Ming Cheng, Kai Zhao, Xin-Yu Zhang, Ming-Hsuan Yang, and
  Philip~HS Torr.
\newblock Res2net: A new multi-scale backbone architecture.
\newblock \emph{TPAMI}, 2019.

\bibitem[Hu et~al.(2018)Hu, Shen, and Sun]{hu2018squeeze}
Jie Hu, Li~Shen, and Gang Sun.
\newblock Squeeze-and-excitation networks.
\newblock In \emph{Proceedings of the IEEE conference on computer vision and
  pattern recognition}, pages 7132--7141, 2018.

\bibitem[Kay et~al.(2017)Kay, Carreira, Simonyan, Zhang, Hillier,
  Vijayanarasimhan, Viola, Green, Back, Natsev, et~al.]{kinetics}
Will Kay, Joao Carreira, Karen Simonyan, Brian Zhang, Chloe Hillier, Sudheendra
  Vijayanarasimhan, Fabio Viola, Tim Green, Trevor Back, Paul Natsev, et~al.
\newblock The kinetics human action video dataset.
\newblock \emph{arXiv preprint arXiv:1705.06950}, 2017.

\bibitem[Ke et~al.(2017)Ke, Bennamoun, An, Sohel, and Boussaid]{ke2017new}
Qiuhong Ke, Mohammed Bennamoun, Senjian An, Ferdous Sohel, and Farid Boussaid.
\newblock A new representation of skeleton sequences for 3d action recognition.
\newblock In \emph{CVPR}, 2017.

\bibitem[Kim and Reiter(2017)]{kim2017interpretable}
Tae~Soo Kim and Austin Reiter.
\newblock Interpretable 3d human action analysis with temporal convolutional
  networks.
\newblock In \emph{CVPRW}, 2017.

\bibitem[Kipf and Welling(2016)]{kipf2016semi}
Thomas~N Kipf and Max Welling.
\newblock Semi-supervised classification with graph convolutional networks.
\newblock \emph{arXiv preprint arXiv:1609.02907}, 2016.

\bibitem[Li et~al.(2019{\natexlab{a}})Li, Li, Zhang, and Wu]{li2019spatio}
Bin Li, Xi~Li, Zhongfei Zhang, and Fei Wu.
\newblock Spatio-temporal graph routing for skeleton-based action recognition.
\newblock In \emph{AAAI}, 2019{\natexlab{a}}.

\bibitem[Li et~al.(2019{\natexlab{b}})Li, Chen, Chen, Zhang, Wang, and
  Tian]{li2019actional}
Maosen Li, Siheng Chen, Xu~Chen, Ya~Zhang, Yanfeng Wang, and Qi~Tian.
\newblock Actional-structural graph convolutional networks for skeleton-based
  action recognition.
\newblock In \emph{CVPR}, 2019{\natexlab{b}}.

\bibitem[Liu et~al.(2016)Liu, Shahroudy, Xu, and Wang]{liu2016spatio}
Jun Liu, Amir Shahroudy, Dong Xu, and Gang Wang.
\newblock Spatio-temporal lstm with trust gates for 3d human action
  recognition.
\newblock In \emph{ECCV}, 2016.

\bibitem[Liu et~al.(2019)Liu, Shahroudy, Perez, Wang, Duan, and
  Chichung]{ntu_120}
Jun Liu, Amir Shahroudy, Mauricio~Lisboa Perez, Gang Wang, Ling-Yu Duan, and
  Alex~Kot Chichung.
\newblock Ntu rgb+ d 120: A large-scale benchmark for 3d human activity
  understanding.
\newblock \emph{TPAMI}, 2019.

\bibitem[Liu et~al.(2020)Liu, Zhang, Chen, Wang, and Ouyang]{msg3d}
Ziyu Liu, Hongwen Zhang, Zhenghao Chen, Zhiyong Wang, and Wanli Ouyang.
\newblock Disentangling and unifying graph convolutions for skeleton-based
  action recognition.
\newblock In \emph{CVPR}, 2020.

\bibitem[Peng et~al.(2020)Peng, Hong, Chen, and Zhao]{peng2020learning}
Wei Peng, Xiaopeng Hong, Haoyu Chen, and Guoying Zhao.
\newblock Learning graph convolutional network for skeleton-based human action
  recognition by neural searching.
\newblock In \emph{in AAAI}, 2020.

\bibitem[Shahroudy et~al.(2016)Shahroudy, Liu, Ng, and Wang]{ntu_60}
Amir Shahroudy, Jun Liu, Tian-Tsong Ng, and Gang Wang.
\newblock Ntu rgb+ d: A large scale dataset for 3d human activity analysis.
\newblock In \emph{CVPR}, 2016.

\bibitem[Shi et~al.(2019{\natexlab{a}})Shi, Zhang, Cheng, and Lu]{ms-aagcn}
Lei Shi, Yifan Zhang, Jian Cheng, and Hanqing Lu.
\newblock {Skeleton-Based Action Recognition with Multi-Stream Adaptive Graph
  Convolutional Networks}.
\newblock \emph{TIP}, 2019{\natexlab{a}}.
\newblock ISSN 10636919.
\newblock \doi{10.1109/CVPR.2019.01230}.

\bibitem[Shi et~al.(2019{\natexlab{b}})Shi, Zhang, Cheng, and
  Lu]{shi2019skeleton}
Lei Shi, Yifan Zhang, Jian Cheng, and Hanqing Lu.
\newblock Skeleton-based action recognition with directed graph neural
  networks.
\newblock In \emph{CVPR}, 2019{\natexlab{b}}.

\bibitem[Shi et~al.(2019{\natexlab{c}})Shi, Zhang, Cheng, and Lu]{shi2019two}
Lei Shi, Yifan Zhang, Jian Cheng, and Hanqing Lu.
\newblock Two-stream adaptive graph convolutional networks for skeleton-based
  action recognition.
\newblock In \emph{CVPR}, 2019{\natexlab{c}}.

\bibitem[Si et~al.(2019)Si, Chen, Wang, Wang, and Tan]{si2019attention}
Chenyang Si, Wentao Chen, Wei Wang, Liang Wang, and Tieniu Tan.
\newblock An attention enhanced graph convolutional lstm network for
  skeleton-based action recognition.
\newblock In \emph{CVPR}, 2019.

\bibitem[Song et~al.(2020)Song, Zhang, Shan, and Wang]{ResGCN}
Yi-Fan Song, Zhang Zhang, Caifeng Shan, and Liang Wang.
\newblock {Stronger, Faster and More Explainable: A Graph Convolutional
  Baseline for Skeleton-based Action Recognition}.
\newblock 2020.
\newblock \doi{10.1145/3394171.3413802}.

\bibitem[Wang et~al.(2020)Wang, Wu, Zhu, Li, Zuo, and Hu]{eca}
Qilong Wang, Banggu Wu, Pengfei Zhu, Peihua Li, Wangmeng Zuo, and Qinghua Hu.
\newblock Eca-net: Efficient channel attention for deep convolutional neural
  networks.
\newblock In \emph{CVPR}, 2020.

\bibitem[Wen et~al.(2019)Wen, Gao, Fu, Zhang, and Xia]{wen2019graph}
Yu-Hui Wen, Lin Gao, Hongbo Fu, Fang-Lue Zhang, and Shihong Xia.
\newblock Graph cnns with motif and variable temporal block for skeleton-based
  action recognition.
\newblock In \emph{AAAI}, 2019.

\bibitem[Yan et~al.(2018)Yan, Xiong, and Lin]{st_gcn}
Sijie Yan, Yuanjun Xiong, and Dahua Lin.
\newblock Spatial temporal graph convolutional networks for skeleton-based
  action recognition.
\newblock In \emph{AAAI}, 2018.

\bibitem[Zhang et~al.(2020)Zhang, Wu, Zhang, Zhu, Lin, Zhang, Sun, He, Mueller,
  Manmatha, et~al.]{zhang2020resnest}
Hang Zhang, Chongruo Wu, Zhongyue Zhang, Yi~Zhu, Haibin Lin, Zhi Zhang, Yue
  Sun, Tong He, Jonas Mueller, R~Manmatha, et~al.
\newblock Resnest: Split-attention networks.
\newblock \emph{arXiv preprint arXiv:2004.08955}, 2020.

\bibitem[Zhang et~al.(2017)Zhang, Lan, Xing, Zeng, Xue, and
  Zheng]{zhang2017view}
Pengfei Zhang, Cuiling Lan, Junliang Xing, Wenjun Zeng, Jianru Xue, and Nanning
  Zheng.
\newblock View adaptive recurrent neural networks for high performance human
  action recognition from skeleton data.
\newblock In \emph{ICCV}, 2017.

\bibitem[Zhang(2012)]{zhang2012microsoft}
Zhengyou Zhang.
\newblock Microsoft kinect sensor and its effect.
\newblock \emph{ACMMM}, 2012.

\end{thebibliography}
\bibstyle{bmvc2k}
\end{document}